\pdfoutput=1

\documentclass[11pt]{article}

\usepackage[final]{acl}

\usepackage{times}
\usepackage{latexsym}

\usepackage{array,float}
\usepackage{listings}
\usepackage{arydshln}
\usepackage[T1]{fontenc}

\usepackage[utf8]{inputenc}

\usepackage{microtype}

\usepackage{inconsolata}

\usepackage{graphicx}

\newcolumntype{M}[1]{>{\centering\arraybackslash}m{#1}}
\newcolumntype{L}[1]{>{\raggedright\arraybackslash}m{#1}}
\usepackage{booktabs}

\newcommand{\treelogo}{\raisebox{5pt}{\includegraphics[scale=0.050]{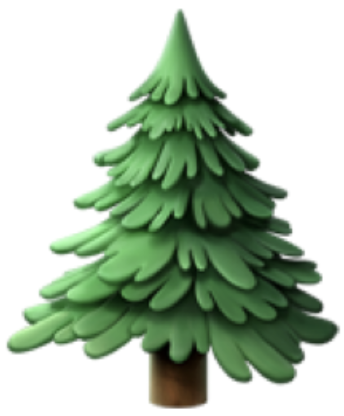}}}
\newcommand{\gtlogo}{\raisebox{3.4pt}{\includegraphics[scale=0.025]{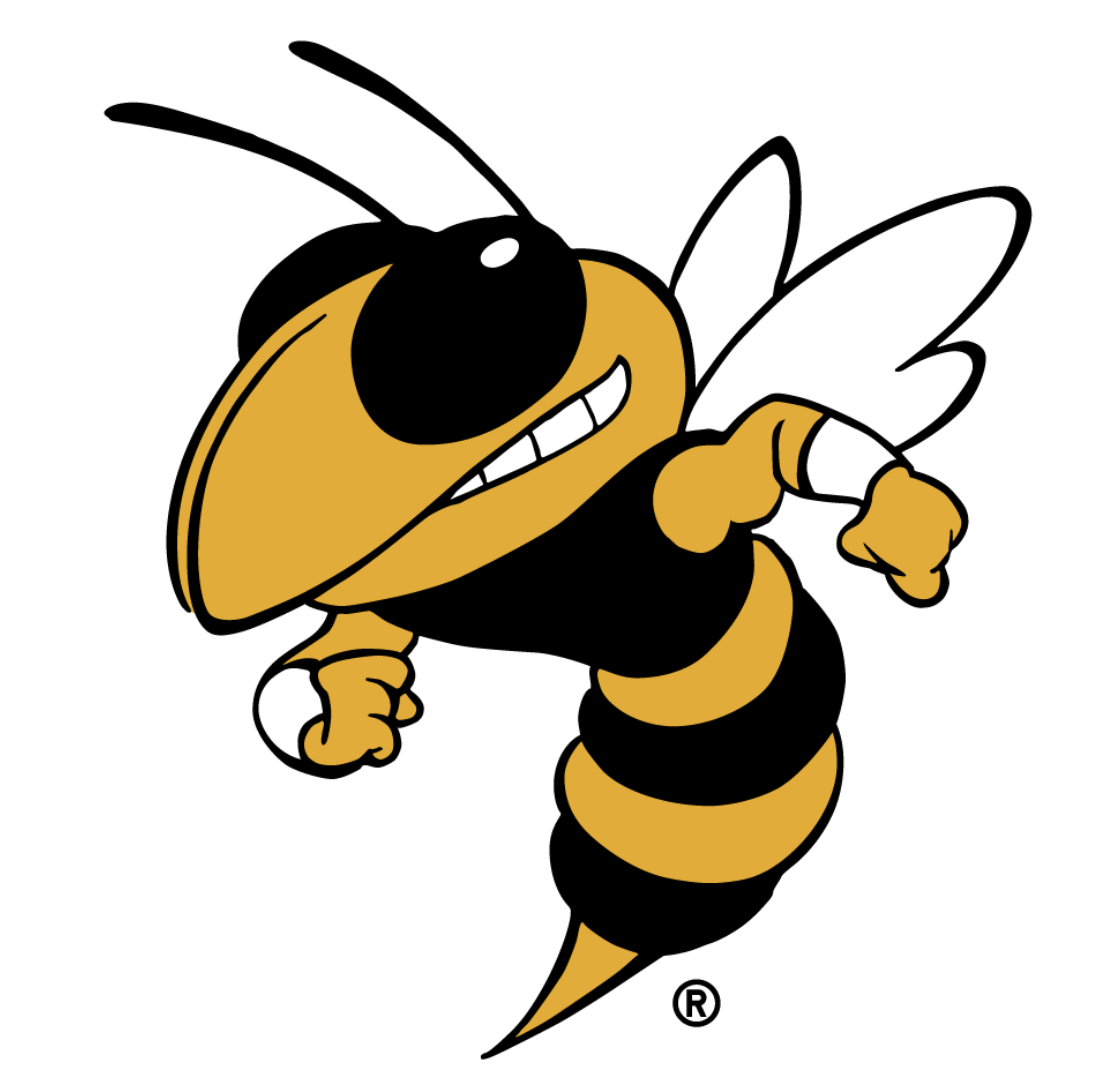}}}
\newcommand{\gt}{\gtlogo}
\newcommand{\stanf}{\treelogo}
\newcommand\coauth{$^\star$}

\title{Social Skill Training with Large Language Models}

\author{Diyi Yang \coauth{} \stanf{} \hspace{1.5em} Caleb Ziems \coauth{} \stanf{}  \hspace{1.5em} William Held \coauth{} \gt{} \hspace{1.5em}\\\textbf{Omar Shaikh} \coauth{} \stanf{} \hspace{1.5em}\textbf{Michael S. Bernstein} \stanf{} \hspace{1.5em} \textbf{John Mitchell}\stanf{} \hspace{1.5em}\\
\stanf{} Stanford University \hspace{1.5em} \gt{} Georgia Institute of Technology\\
\texttt{\href{mailto://diyiy@cs.stanford.edu}{diyiy@cs.stanford.edu}} \hspace{1.5em} \texttt{\href{mailto://cziems@stanford.edu}{cziems@stanford.edu}} \hspace{1.5em}\texttt{\href{mailto://wheld3@gatech.edu}{wheld3@gatech.edu}} \\\hspace{1.5em} 
\texttt{\href{mailto://oshaikh@stanford.edu}{oshaikh@stanford.edu}} \hspace{1.5em} \texttt{\href{mailto://msb@cs.stanford.edu}{msb@cs.stanford.edu}} \hspace{1.5em} \texttt{\href{mailto://mitchell@cs.stanford.edu}{mitchell@cs.stanford.edu}}
}

\newcommand\blfootnote[1]{%
  \begingroup
  \renewcommand\thefootnote{}\footnote{#1}%
  \addtocounter{footnote}{-1}%
  \endgroup
}

\begin{document}
\maketitle
\begin{abstract}
People rely on social skills like conflict resolution to communicate effectively and to thrive in both work and personal life. However, practice environments for social skills are typically out of reach for most people. How can we make social skill training more available, accessible, and inviting? Drawing upon interdisciplinary research from communication and psychology, this perspective paper identifies social skill barriers to enter specialized fields. Then we present a solution that leverages large language models for social skill training via a generic framework. Our \textbf{AI Partner, AI Mentor} framework merges experiential learning with realistic practice and tailored feedback. This work ultimately calls for cross-disciplinary innovation to address the broader implications for workforce development and social equality. \blfootnote{\coauth Equal contribution.}  
\end{abstract}

\section{Introduction}
People need both general and domain-specific skills to succeed in home and work life \cite{dean2017soft}. Specialized workers need not only technical competence, but also field-specific soft skills that extend broader social skill-sets. For example, mental health counselors use active listening \cite{nemec2017can}, a skill for building trust and empathy \cite{devito2019interpersonal,ramsey1997listening}. Similarly, business managers employ conflict resolution \cite{de2001theory} and strengthen team bonds \cite{devito2019interpersonal} with specialized strategies \cite{lipsky2003emerging}. Learning these skills may involve passive observation, trial-and-error, or explicit instruction, but ultimately, a learner will need deliberate practice \cite{giddens2006sociology}, as social skills are inherently interactive.

Learning environments for social skills can be inaccessible, especially when training is offered by experts in formal programs, which are expensive, time-consuming, and limited in availability. Existing mechanisms for practice and feedback largely rely on expert supervision, making training difficult to scale. 
In particular, there may be a shortage of professionally trained coaches \cite{hoffmann2023association,wiggan2021national}, and most coaches who can provide tailored feedback are not able to help the large number of people who need it. 

Practicing with peers can be a viable alternative only if peers are experienced. Individuals may also find it challenging or unsafe to practice high-risk tasks. Many people, especially from underrepresented groups and disadvantaged populations, have limited opportunities and social capital to learn and practice their target field's specialized skill frameworks, which can exacerbate social inequality \cite{ovink2011more}. We argue that large language models can help make social skill training more accessible, safe, and inviting, with tailored feedback in realistic, virtual practice spaces.

In this position paper, we propose \textbf{Social Skill Training} via two complementary visions of AI assistance, and outline a roadmap for their implementation. The first vision is that of the AI Partner, which can provide a scalable solution to experiential learning through simulated practice. Already, research has shown that human role-play can effectively teach communication, cooperation, and leadership skills \cite{gjeraa2014efficacy,deutsch2011handbook}. Compared to on-the-job training, simulations allow learners to assume fewer risks and opportunity costs. By making simulations accessible, the AI Partner will reduce the socioeconomic barrier to enter specialized fields. Our complementary vision is the AI Mentor, which will offer personalized feedback based on domain expertise and factual knowledge. Together, the AI Partner and AI Mentor (APAM) framework can merge experiential learning with realistic practice and tailored feedback. Our paper calls for cross-disciplinary innovation to address APAMs broad implications.

\section{LLMs for Characters and Simulation}
Prior research has shown simulation-based learning to be a highly effective educational tool~\cite{blair2007pedagogical,tambe1995intelligent,chernikova2020simulation}. These studies worked with manually constructed simulation templates. In this work, we will focus on LLM-based simulations as a more flexible and scalable solution. 

Prompted LLMs can effectively roleplay believable characters
\cite{argyle2023leveraging,park2022social}, who operate in specific contexts with plausible behaviors \cite{park2023generative}, realistic preferences \cite{horton2023large} and decision-making strategies \cite{zhao2024expel}, human-like negotiation tactics \cite{gandhi2023strategic}, and empirically-attested psychological responses \cite{aher2023using}. Agent-based simulations, powered by LLMs, have been used for understanding debates \cite{du2023improving}, strategic communication \cite{xu2023exploring}, collaboration \cite{zhang2023exploring}, conflict \cite{hua2023war}, online behavior \cite{ren2024bases}, and even urban planning \cite{zhou2024large}.

Towards AI Partner design, the set of methodologies is rapidly expanding. Prompting can further be used alongside reinforcement learning to update LLMs according to a set of high-level guiding principles called a constitution~\cite{bai2022constitutional}, which can be written in plain text for rapid prototyping \cite{petridis2023constitutionmaker}. Extensions of LLM-based dialogue models include architectures for character consistency~\cite{touvron2023llama}, and fine-tuning on datasets specific to character simulation and conversation 
\cite{thoppilan2022lamda,shuster2022blenderbot,kwon2023and}.

The development of the AI Mentor requires special care. Unlike the AI Partner, which may simulate real-world misconceptions, the AI Mentor should stay grounded in recent expert knowledge, which may not be present in the model's original training corpus. As a possible solution, Retrieval Augmented Generation \cite[RAG;][]{lewis2020retrieval} can fetch relevant knowledge from external sources and dynamically update the prompt
\cite{xu2021beyond,shuster2022language,jiang2023active,khattab2022demonstrate}. The use of these approaches largely aids the process of leveraging knowledge from textbooks and other curated sources. 

The APAM framework may employ LLMs as only one part of a larger system that integrates retrieval, tool use~\cite{schick2024toolformer}, constitutional decision making, and generation into a single pipeline \cite{wang2024survey,wu2022promptchainer,yang2022re3} via planning modules \cite{shinn2023reflexion,yao2022react}. Planning can rely on traditional search algorithms \cite{yao2024tree}, or may separately prompt another LLM to evaluate, filter, and rank candidate plans\cite{shinn2023reflexion,yao2022react}. Planning may even rely on LLM-generated code, which is then executed to filter viable candidates~\cite{wang2023voyager}.

\begin{figure*}
    \centering
    \includegraphics[width=1\textwidth]{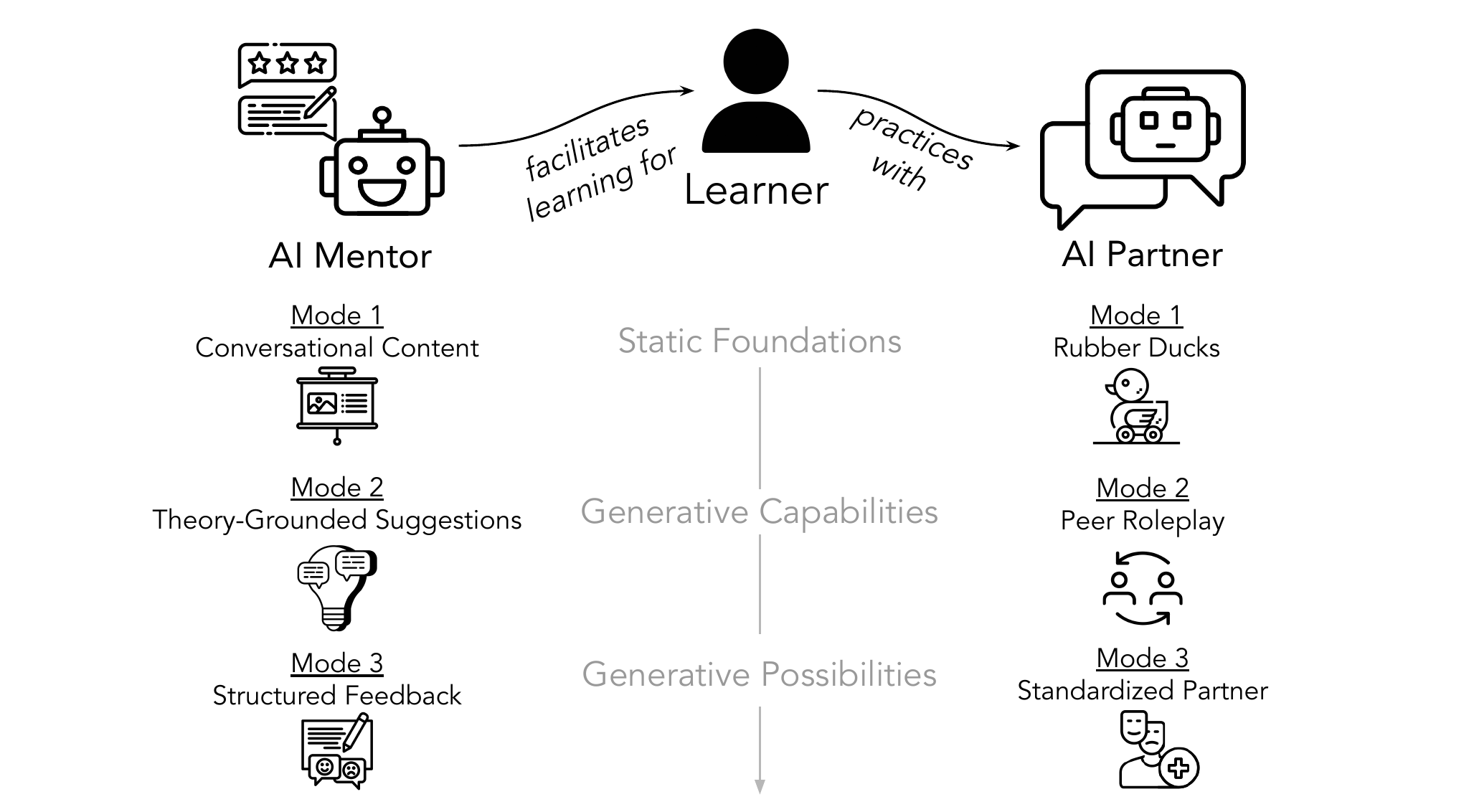}
    \caption{\textbf{Modes of the APAM framework.} As AI capabilities improve, the APAM framework develops from its basis in non-AI teaching practices towards the possibility of realistic simulated AI Partner learning scenarios augmented with  AI Mentor feedback that can be personalized based on prior practice sessions between the User and the AI Partner. With LLMs, our prior work has shown that AI Mentors can effectively generate suggestions based on best practices \cite{hsu2023helping} and AI Partners can replicate many of the benefits of roleplay \cite{shaikh2023rehearsal}.}
    \label{fig:enter-label}
\end{figure*}

\section{The APAM Framework}
We propose a generic framework for social skill training with an \textbf{AI Partner} and an \textbf{AI Mentor} (APAM). Both are critical. When a user wants to learn a new social skill, the AI Partner can help them practice a relevant scenario with simulated conversation. The AI Mentor can provide knowledge-grounded feedback at critical junctures of the simulation.

\subsection{AI Partner} Constructing and deploying an AI Partner is non-trivial for multiple reasons. First, it is difficult to maintain \textbf{consistency} in the stylistic, behavioral, and emotional characteristics of the simulated partner~\cite{weston2023system}. Second, \textbf{faithful} simulations require a high level of complexity and detail that align with the target domain. Third, simulations should follow an \textbf{efficient curriculum} that quickly and deeply educates the learner. Thus, the AI partner should exhibit a \emph{consistent, plausible, and instructive personality}. Note that \textbf{diversity} is one key component in an instructive system, which requires the simulated AI partner to demonstrate diverse social and cultural attributes. Through these dimensions, LLMs offer an actionable path to realize the ideal AI Partner. 

\subsection{AI Mentor} The development of AI Mentor heavily relies on domain expertise, context awareness, and feedback efficacy, in addition to consistency that AI partners exhibit. \textbf{Domain expertise} means the AI Mentor should cite and elaborate on theories or frameworks from the related literature to develop their feedback, such as psychotherapy, conflict resolution, or negotiation, rather than producing generic and broad responses. \textbf{Context awareness} means that the AI Mentor should ground its suggestions in both the current scenario and the learner's knowledge state, rather than citing generic or random facts. This creates technical challenges surrounding the handling of long context and heterogeneous input. \textbf{Feedback efficacy} means the mentor should personalize the communicative style, timing, specificity, and granularity of its feedback to most effectively empower the learner at a given moment. Feedback should also be empathetic, respectful, and appropriate to the cultural and social context. 

\subsection{Methodology}
\begin{table*}[]
\resizebox{\textwidth}{!}{
\renewcommand{\arraystretch}{1.25}
\begin{tabular}{cccccc}
\hline
& \multicolumn{5}{c}{\textbf{Social Skill Clusters}}
\\
\cmidrule{2-6}
& \textbf{\begin{tabular}[c]{@{}c@{}}Active Listening\\ \tiny{\cite{rogers1957active}}\end{tabular}} & \textbf{\begin{tabular}[c]{@{}c@{}}Conflict Avoidance
\\\tiny{\cite{morris2006teachers}}
\end{tabular}} & \textbf{\begin{tabular}[c]{@{}c@{}}Conflict Resolution\\\cite{behfar2008critical}\end{tabular}} & \textbf{\begin{tabular}[c]{@{}c@{}}Empathy \\\cite{smith2006cognitive}\end{tabular}} & \textbf{\begin{tabular}[c]{@{}c@{}}Rhetoric\\\cite{AristotleRhetoric}\end{tabular}} \\ \hline
\begin{tabular}[c]{@{}c@{}}\textbf{Description}\end{tabular} & 
\begin{tabular}[c]{@{}c@{}}Listening to express\\understanding of the \\speaker’s  intentions.\end{tabular} & \begin{tabular}[c]{@{}c@{}}The ability to\\ prevent disagreements \\or differences\\of opinion.\end{tabular} & \begin{tabular}[c]{@{}c@{}}The ability to resolve\\ disagreements or \\differences of opinion\end{tabular} & \begin{tabular}[c]{@{}c@{}}The ability\\ to understand\\another person's\\ experience\end{tabular} & \begin{tabular}[c]{@{}c@{}}The ability to\\present strong arguments\\for one's beliefs\end{tabular}\\ \hline
\textbf{Evaluation} & \begin{tabular}[c]{@{}c@{}}Active Listening\\Attitude Scale\\ \cite[][]{mishima2000development}\end{tabular} & \begin{tabular}[c]{@{}c@{}}Dutch Test\\for Conflict Handling\\\cite[][]{van2013complex}\end{tabular} & \begin{tabular}[c]{@{}c@{}}Dutch Test\\for Conflict Handling\\\cite[][]{van2013complex}\end{tabular} & \begin{tabular}[c]{@{}c@{}}Jefferson Scale\\\cite{hojat2001jefferson}\end{tabular} & \begin{tabular}[c]{@{}c@{}}Facilitative\\ Interpersonal Skills\\\cite[][]{anderson2007facilitative}\end{tabular}\\ \hline
\begin{tabular}[c]{@{}c@{}}\textbf{Application}\\\textbf{Domain}\end{tabular} & 
\begin{tabular}[c]{@{}c@{}}Counseling\\ \cite{nemec2017can}\end{tabular} & \begin{tabular}[c]{@{}c@{}}Classroom\\ Management\end{tabular} & \begin{tabular}[c]{@{}c@{}}Product Management\\{\cite{lipsky2003emerging}}\end{tabular} & \begin{tabular}[c]{@{}c@{}}Nursing\\\tiny{\cite{yu2009evaluation}}\end{tabular} & \begin{tabular}[c]{@{}c@{}}Litigation\\ \cite{singer1988persuasion}\end{tabular}\\ \hline
\begin{tabular}[c]{@{}c@{}}\textbf{Domain-}\\\textbf{Specific}\\\textbf{Framework}\end{tabular} & 
\begin{tabular}[c]{@{}c@{}}Motivational\\ Interviewing\\ \cite{moyers2014motivational}\end{tabular} & \begin{tabular}[c]{@{}c@{}}Positive Behavioral\\ Interventions\\and Supports\\\cite[][]{bradshaw2012effects}\end{tabular} & \begin{tabular}[c]{@{}c@{}}Alternative Dispute\\Resolution \\\cite[][]{lipsky2003emerging}\end{tabular} & \begin{tabular}[c]{@{}c@{}}Person-Centered\\Nursing \\\tiny{\cite{mccormack2006development}}
\end{tabular} & \begin{tabular}[c]{@{}c@{}}CREAC Legal\\Writing Paradigm \\\cite{kraft2014creac}\end{tabular}\\ \hline
\begin{tabular}[c]{@{}c@{}}\textbf{Learner}\end{tabular} & 
\begin{tabular}[c]{@{}c@{}}Novice Therapist\end{tabular} & \begin{tabular}[c]{@{}c@{}}Teacher-in-Training\end{tabular} & \begin{tabular}[c]{@{}c@{}}Manager\end{tabular} & \begin{tabular}[c]{@{}c@{}}Nurse-in-Training\end{tabular} & \begin{tabular}[c]{@{}c@{}}Novice Litigator\end{tabular}\\ \hline
\begin{tabular}[c]{@{}c@{}}\textbf{AI Partner}\end{tabular} & 
\begin{tabular}[c]{@{}c@{}}Digitized Patient\end{tabular} & \begin{tabular}[c]{@{}c@{}}Virtual Student\end{tabular} & \begin{tabular}[c]{@{}c@{}}Simulated Dispute\end{tabular} & \begin{tabular}[c]{@{}c@{}}Digitized Patient\end{tabular} & \begin{tabular}[c]{@{}c@{}}Simulated Courtroom\end{tabular}\\ \hline
\begin{tabular}[c]{@{}c@{}}\textbf{AI Mentor}\end{tabular} & 
\begin{tabular}[c]{@{}c@{}}Expert Counselor\end{tabular} & \begin{tabular}[c]{@{}c@{}}Experienced Teacher\end{tabular} & \begin{tabular}[c]{@{}c@{}}Mediator\end{tabular} & \begin{tabular}[c]{@{}c@{}}Experienced Nurse\end{tabular} & \begin{tabular}[c]{@{}c@{}}Expert Lawyer\end{tabular}\\ \hline
\end{tabular}
}
\caption{\textbf{Different use cases of APAM framework.} Therapists and other specialists depend on general skill clusters like \textit{active listening}, which are formalized in domain-specific frameworks like \textit{motivational interviewing}. In this example \textit{(left column)}, the AI partner might be a \textit{digitized patient}, while the AI mentor is an \textit{expert counselor}.}
\label{tab:use_cases}
\end{table*}

We now propose a generic methodology for Social Skill Training via LLMs in four steps: (i) understanding the social processes that underlie one’s desired skill (e.g., conflict resolution); (ii) designing an AI partner to simulate conversations that expose the learner to the target processes, allowing the learner to practice; (iii) creating an AI mentor to provide tailored feedback; (iv) integrating the two agents into a simulated environment for users to learn safely. These four steps ensure effective social skill training. It is only through simulation in Step ii that users can practice realistically, and domain knowledge in Step iii that the system can provide pedagogically effective feedback. Finally, we can determine the success of our system in Step iv when we run comparative user studies. 

Beginners are the ideal audience for the APAM framework, but skilled workers could also use APAM systems to refresh their knowledge. Even if AI partners have imperfections in simulation and AI mentors provide relatively rigid theoretical feedback, the APAM framework can provide benefits by structurally facilitating exploration and analytical self-reflection (e.g., rubber duck debugging)~\cite{schon1986reflective, ku2010metacognitive}. APAM focuses on empowering users to become more aware of where they struggle.

\subsection{Examples of APAM}
\label{subsec:apam_examples}
There are many domains where APAM can improve learners’ skills. Table~\ref{tab:use_cases} samples five broad skill clusters (e.g. \textit{active listening}) which translate into career-specific domains (\textit{mental health counseling}) through field-specific frameworks (\textit{motivational interviewing}). These broad skill examples come with highly-developed psychological tests and self-report inventories to assess learning objectives. Our framework is not limited only to such canonical examples; we will discuss how to evaluate APAM systems more broadly in \S\ref{sec:eval}.

Recent work already exemplifies the APAM framework. For instance, \citet{sharma2023human} developed Hailey, an AI Mentor that provides feedback to mental health supporters when writing empathetic messages. To promote better political engagement online, \citet{argyle2023leveraging} developed an AI Mentor system that can provide feedback on polite and validating discourse strategies.
In the legal domain, \citet{jiang2024leveraging} leveraged LLMs to help non-experts learn intricate legal concepts to make legal knowledge accessible for encouraging civic participation in democracy. Besides these examples, we now discuss three APAM applications in more detail: CARE for peer counseling, Rehearsal for conflict resolution, and GPTeach for education. 

\paragraph{CARE (AI Mentor)} Peer counseling platforms depend on effective volunteer counselors, but most volunteers do not have access to personalized learning resources. One of our recent works introduces CARE: an interactive AI Mentor that trains peer counselors with automatic suggestions \cite{hsu2023helping}. During the practical training stage, CARE diagnoses which counseling strategies are most suitable in the given context and suggests tailored responses. Counselors can choose to select, modify, or ignore these suggestions. We find that this LLM-based system, trained on Motivational Interviewing strategies from counseling conversation data, significantly helps novice counselors respond to challenging situations. 

\paragraph{Rehearsal (AI Partner)}  Conflict is an uncomfortable and unavoidable part of life, but people can learn conflict resolution through deliberate and strategic practice. Since few people have access to the necessary training resources, we developed the Rehearsal \cite{shaikh2023rehearsal} system to provide these at scale. Rehearsal helps users practice conflicts with a believable simulated interlocutor (AI Partner), identify alternative conversational paths, and learn through feedback on how to apply specific conflict strategies (AI Mentor). With Rehearsal, users can practice predefined conflict scenarios, or define their own novel scenarios. Our between-subjects evaluation showed that Rehearsal significantly helps learners navigate later unaided conflict compared to control groups. 

\paragraph{GPTeach (AI Partner)}  For a teaching assistant (TA), three important domain-specific skills are academic communication, classroom management, and pedagogy. Learning these skills on-the-job can be stressful \cite{eddy2013new}. However, novice TAs in particular rarely have the resources they need to develop these skills before entering the classroom. TA instruction is often limited to static guides written by expert TAs; the first time new TAs ever practice is with actual students. To reduce potential harms, TAs should have a space to practice their teaching skills beforehand. To this end, GPTeach \cite{markel2023gpteach} uses LLMs to simulate a programming course in which simulated students make mistakes and ask questions like real students. This allows novice TAs to practice across a wide range of student behaviors.

\section{Vision for Safe Deployment}
LLMs have strong potential as tools for social skill training because they can flexibly generate coherent and natural text without the need for extensive topic-specific engineering used by prior works \cite{biswas2005learning}. However, this flexibility often comes with more limited controllability, making such deployment dangerous for high-risk scenarios like therapy or mental health crises. 

Our APAM framework provides guidelines for how to safely use AI in social skill training by decomposing safe usage into a continuum. In this section, each safety recommendation is tailored to a specific level of system capabilities. The different modes below represent different capability clusters that one might foresee from AI Mentors and AI Partners. By selecting a mode dependent on current capabilities and limitations, one can safely deploy valuable LLMs without requiring solutions to difficult open technical safety questions.

\subsection{AI Partner Continuum}
\paragraph{Mode 1: Rubber Ducking}
Simulation-based learning is grounded in a wealth of cross-disciplinary research 
\cite{cherryholmes1966some,dorn1989simulation,randel1992effectiveness,kincaid2003simulation,brennan2013effects}. Even simple, low-fidelity simulations can prove effective, and to demonstrate this, we will consider the least developed partner: a passive, inanimate object. The practice of explaining your ideas to a rubber duck is called ``Rubber ducking.” Despite being a passive “partner,” the rubber duck helps learners identify mistakes through the power of social learning and explanation \cite{ku2010metacognitive}. While today’s LLMs are certainly more powerful than a rubber duck, this highlights how “partners” can be valuable and safe even without human-level capabilities.

\paragraph{Mode 2: Peer Roleplay}
Current Partner technologies (e.g., Rehearsal) resemble untrained peers roleplaying unfamiliar situations. While these simulations share surface level characteristics with real social situations, they often lack nuance, especially for roles which peers may not have lived experience with~\citep{matz2010using}. Despite this shortcoming, roleplay has long been a valuable tool for curriculum design, since it can help move learners from abstract theories to real-world practice.
 
\paragraph{Mode  3: Standardized Partner}
In high-risk domains like therapy, AI Partners will need to maintain a higher standard of consistency and reproducibility than most LLM-based simulation systems. We call this higher standard the Standardized Partner, like the “\emph{Standardized Patients}” from medical training \cite{van1990assessment} who are professionals trained to reproducibly simulate a patient with specific personality traits and ailments. In Medicine, Standardized Patients can prepare students as effectively as expert practitioners which shows that expert-training AI may not require the development of expert AI. Achieving this requires addressing the stereotypes \cite{shaikh2022second}, caricature and tropes \cite{cheng2023marked,cheng2023compost}, as well as misinformation \cite{lin2021truthfulqa} produced by today’s LLMs.

\subsection{AI Mentor Continuum}
\paragraph{Mode 1: Conversational Content}
Where AI Partners help learners learn through experience, AI Mentors connect formal or theoretical knowledge to these experiences. Fundamentally, Mentors can also be grounded in non-AI teaching principles: when educational materials follow a conversational rather than formal style, learning outcomes improve consistently \cite{sorden2012cognitive}. The simplest AI Mentors are systems that rephrase dense or distractingly formal academic texts into the most understandable register. 

\paragraph{Mode  2: Theory-Grounded Suggestions}
Instead of presenting theories in the abstract, systems can offer theory-grounded suggestions to remind learners of the expert theories more naturally. Importantly, the suggestion format does not require that the system has perfect knowledge, since learners can benefit from judging even imperfect suggestions, developing their own intuitions about when theory applies. CARE \cite{hsu2023helping} is one such work that tests the limits of these capabilities.

\paragraph{Mode  3: Structured Feedback}
AI Mentors can be improved to provide structured, actionable, and personalized suggestions with a greater scope and time-scale than the local-level feedback of Mode 2. This would require reasoning over long, multi-turn conversations to an extent not possible with the attention mechanisms and context length limitations of current LLMs \cite{liu2024lost}. The technical requirements of such an AI Mentor may be greater than that of developing an AI Expert directly, as teaching a task can be more challenging than performing a task. We believe this challenge is merited as AI Mentors can create lasting value even after the exposure to the AI Mentor system ends. This enhances rather than diminishes human expertise in critical areas and avoids creating an ongoing dependency on AI. 

\section{Technical Challenges}
To create effective avenues for social skill training, APAM-systems require work on concrete technical challenges. In this section, we outline a core subset, prioritizing long-term interaction, expert-driven design of APAM systems, personalization, and designing for end-user interaction. 

\subsection{Optimizing Long-Term Interactions}
\paragraph{Persona Consistency} First, LLMs should remain consistent when simulating social skill training. When individuals practice with LLMs over multiple turns, an LLM should not "forget" aspects in the initial prompt, e.g. by providing feedback unrelated to the initial instruction or ignoring attributes of the roleplayed simulation. Like~\citet{weston2023system}, we suspect a limitation of the attention mechanism in LLMs. As the context window increases, models may place less attention on initial instructions. One avenue of future work involves designing modeling or prompting methods to enforce consistency. Already, Ghost Attention~\cite{touvron2023llama} and System Two Attention~\cite{weston2023system} offer technical solutions to maintaining instruction consistency. Beyond modeling solutions, benchmarking consistency across multi-turn skill training---either by collecting real-world interaction datasets or constructing a static benchmark---would highlight deficiencies, addressable by future work.  

\paragraph{Conversational Grounding} is a fundamental component of interpersonal interaction~\cite{clark1996using}. We often take multiple turns to establish an utterance as common ground. Humans do not simply "follow instructions"---we ask clarification questions, follow up on underlying details, or repair assumptions made by other interlocutors~\cite{clark1989contributing}. This is critical for social skill training: dialogue agents must \textit{build grounding} with individuals before generating feedback. Current instruction-following LLMs, however, are trained on single-step interactions~\cite{ouyang2022training}. Thus, contemporary LLMs generate text that \textit{assumes} characteristics about an individual~\cite{shaikh2023grounding, chiu2024computational}. In educational or mental health contexts, assuming the source of struggles can result in irrelevant feedback~\cite{graesser1995collaborative, strumwasser1991appropriateness, wang-etal-2024-backtracing}. Integrating multi-turn preference optimization into LLMs is one promising avenue for future work; \citet{hong2023zero} and \citet{andukuri2024star}, for example, explore RLHF and self-improvement methods to generate conversational grounding. However, identifying \textit{where} and \textit{why} humans take time to establish common ground across a diverse set of situations---and training LLMs to reflect this behavior---is still an open problem.

\subsection{Integrating Expert Frameworks}
For training systems to be effective and safe~\citep{demszky2023using}, they should be closely integrated with domain-specific expert skill frameworks like \textit{motivational interviewing} (c.f., Table~\ref{tab:use_cases}). With LLM agents, however, adherence to specific frameworks is not guaranteed. By default, LLMs generate text in an unconstrained fashion. Ensuring generation adherence to expert frameworks is a highly domain-specific process. For effective interaction, experts must first outline and demonstrate specific strategies and constraints~\cite{agrawala2011}. Learning how to integrate these constraints into an LLM---either by finetuning a new model, designing new constrained decoding processes~\cite{keskar2019ctrl}, or building prompting pipelines~\cite{wu2022promptchainer}---are important avenues for future work. Finally, APAM-based systems should allow a practitioner to reflect on theory in structured feedback or peer role-play~\cite{schon1986reflective}.  For example, technical methods should enable users to explore counterfactual roleplays or feedback grounded in theory.   

\subsection{Designing for End-User Control}
While we focus on technical methods that improve social skill training for LLMs, we note that these methods should be amenable to user adjustments. If an individual decides to change the underlying framework used by an LLM for training, adjustments should be directly possible in the system itself. Systems like ConstitutionMaker~\cite{petridis2023constitutionmaker}, for example, allow end-users to design and edit prompting principles through an interactive interface. Similarly, new technical methods should come with interactive complements that enable direct control~\cite{shneiderman1983direct}. Since training systems are inherently user-facing, designing interactive interfaces that allow for accessible control---either by an expert or learner---will allow individuals to customize the type of training they receive, instead of depending on a researcher. 

\subsection{Personalizing Skill Training}
Personalization is a key challenge even for general tasks around LLMs \cite{mysore2023pearl,tan2024democratizing,wu2021personalized}. This connects to the \emph{consistency} and \emph{diversity} attributes of AI Partner, as well as \emph{feedback efficacy} of AI Mentor.
Effective skill training tailors feedback and experiences to meet the needs of each learner. Such personalized training has been made possible via APAM as learners can select and design AI partners or mentors that are relevant to them.   
It is, however, not trivial to operationalize personalization \cite{flek-2020-returning,dudy2021refocusing}. Prior research has investigated various writing styles (e.g., formal vs informal, simplified vs sophisticated language usage) \cite{alhafni2024personalized,li2023automatic}, and learners' expertise in certain topics \cite{bernacki2021systematic}. Building upon these prior studies and taking into account the appropriate set of personalization-related attributes---as well as learners' knowledge or expertise backgrounds~\cite{huang2012empowering}---becomes increasingly important. 

\section{Evaluation}
\label{sec:eval}
\begin{table*}[t]
\centering
\resizebox{\textwidth}{!}{
\begin{tabular}{|M{0.25\textwidth}|M{0.5\textwidth}|M{0.25\textwidth}|M{0.12\textwidth}|M{0.12\textwidth}|}
\hline
\multicolumn{5}{|c|}{\textbf{Intrinsic Evaluation}} \\ \hline
\textbf{Metric Type} & \textbf{Description} & \textbf{Examples} & \textbf{Applicability} & \textbf{Category}\\ \hline
Reference Based Evaluation & Metrics of the similarity and distinguishability between a systems interactions and a set of gold standard interactions.  & ~\citep{hashimoto-etal-2019-unifying} & APAM & Automated \\ \cdashline{1-4}
Topic Analysis       & Assessment of relevance of topics covered compared to expectations. &  ~\citep{cheng2023compost}   & AP &    \\ \cdashline{1-4}
Classifier Based Scoring & Using trained classifiers to categorize the frequency of known effective and realistic behaviours.  & ~\citep{sharma2024investigating} & APAM & \\  \cdashline{1-4}
LLM Prompt Scoring       & Prompting LLMs to act as automated judges which provides Likert scale scores for a simulation.            &  ~\citep{zhou2023sotopia, dubois2024alpacafarm}      & AP &     \\ \hline
Human Ranking & Comparative metric where systems are ranked based on a rubric of evaluation. & ~\citep{park2023generative, zhou2024real}       & APAM &      \\ \cdashline{1-4}
Human Scoring & Likert Scale ratings of the system along given a rubric of evaluation. & ~\citep{thoppilan2022lamda}       &  APAM &   \\ \cdashline{1-4}
Suggestion Usage & Rate at which participants utilize suggestions provided by an AI mentor system. &~\citep{hsu2023helping} & AM & User \\ \cdashline{1-4}
Recommender Score & Rating of how likely a user would be to recommend the system to a friend. & ~\citep{markel2023gpteach} & APAM & \\ \hline
\multicolumn{5}{|c|}{\textbf{Extrinsic Evaluation}} \\ \hline
\textbf{Metric Type} & \textbf{Description} & \textbf{Examples} & \textbf{Applicability} & \textbf{Category}\\ 
\hline

Behavioral Impacts & Changes in qualitatively coded participant behaviors before and after exposure to the system. &~\citep{shaikh2023rehearsal, markel2023gpteach} & APAM & Short-Term \\ \cdashline{1-4}
Self-Efficacy Reports & Changes in participants' self-reported efficacy on the skills practiced before and after exposure. & \citep{shaikh2023rehearsal}    & APAM    &     \\\cdashline{1-4}
Standardized Evaluation       &  Changes in participant scores on closed-ended assessments of knowledge about the skills practiced.           &     \citep{shaikh2023rehearsal}   &  APAM &   \\ \cdashline{1-4}
Short-Term Economic Outcomes & Impacts of a training program on participants short-term wages and employment. & \citep{adhvaryu2018skills, chioda2021making} & APAM  & \\ 
\hline
Non-Financial Benefits & Impacts on non-financial measures such as health, risk-taking behaviors, and levels of societal trust. & \citep{oreopoulos2011priceless, heckman2012hard}  &  APAM  & Long-Term \\ \cdashline{1-4}
Long-Term Economic Outcomes       & Impacts of a training program on long-term earnings, workplace stability, and economic mobility.             &  \citep{barrera2023hard}   &  APAM    &       \\ \hline
\end{tabular}%
}
\caption{\textbf{Intrinsic and Extrinsic Evaluation Procedures applicable to APAM systems from prior work.} At present, Natural Language Processing practitioners primarily focus on intrinsic evaluations for their systems. Here, we stress the importance of evaluating APAM systems using established measures for educational outcomes.}
\label{table:metrics}
\end{table*}

The evaluation of AI partners and AI mentors is a major challenge; tools based on APAM involve complex computational systems and interaction with users with varying desires and backgrounds. To develop these training tools as a field, evaluation measures need to move beyond metrics traditionally used in Natural Language Processing to protocols from multiple relevant fields and stakeholders. Including multidisciplinary perspectives will help evaluate the empirical performance of such systems, their usability from a user perspective, and their long-term impacts on both users and their communities.

At present, research on text generation focuses largely on intrinsic evaluations which assess the quality of outputs with predefined rubrics or interactions. In Table \ref{table:metrics}, we separate these into fully automated evaluations and user-driven evaluations. Reference-based metrics, such as perplexity or Kullback–Leibler divergence~\citep{theis2016note}, are common automated assessments of system quality, as they are both simple and allow for a rich definition of desired behavior through demonstrations. While many works aim to optimize metric reliability~\citep{hashimoto-etal-2019-unifying}, they often lose statistical power as researchers implicitly optimize metrics on fixed datasets~\citep{goyal2023news}.

In pursuit of more flexible assessments, practitioners often turn to human qualitative assessments of systems, either through Likert scale-based scoring or comparative ranking of systems. In these procedures, system developers develop a rubric of desirable qualities such as believability, helpfulness, and consistency. These metrics are a useful gauge of the quality of these interactive systems.

However, as systems become more generally coherent and rubrics become more fine-grained, these methods of human validation often raise reproducibility concerns~\citep{karpinska-etal-2021-perils}. While LLMs themselves are increasingly used to replace the human annotators in these processes~\citep{dubois2024alpacafarm}, this raises separate concerns about the systemic judgment biases of the LLM as a judge~\citep{zheng2024judging}. As such, other studies have focused on more coarse, functional metrics from user studies such as the Recommender Score~\citep{markel2023gpteach} or the rate at which users make use of system outputs~\citep{hsu2023helping}. 

To develop effective evaluations of more powerful systems, we believe domain users need to be involved as collaborators, rather than just as annotators. Potential users are best placed to assess the intrinsic measures that make a system usable, confusing, or even harmful. In current procedures, users are assigned to predefined roles assessing systems along strictly defined rubrics created by the system designers which centers the process on the developer's preconceptions~\cite {birhane2022power}. 

Resolving this is not a simple matter of involving different individuals for the above evaluation metrics. Instead, potential users should be involved as stakeholders in high-level design---before development begins---to center the process around recognizing end-user expertise. For example, involving experts in the design of a training tool may highlight pedagogical theory overlooked by researchers. Watching end-users interact with prototype APAM systems will highlight opportunities for new technical methods or interactions. Practices from participatory design in HCI can serve as helpful guidelines for APAM platforms~\cite{muller1993participatory, schuler1993participatory}. 

Ultimately, however, extrinsic evaluation of how interaction with a tool changes the behavior of the individuals who use it. In the case studies we cover, measurable changes in behavior, self-efficacy reports, and standardized test scores have been used to evaluate short-term impacts.  Whether these shifts are sustained and whether they create beneficial outcomes for participants, however, requires moving beyond what is currently evaluated in NLP. Educators and economists, however, have deep experience designing studies to evaluate the long-term impacts of training tools and education. This is most commonly utilized for economic outcomes, since these are the most straightforwardly measured benefits of soft-skills training~\citep{adhvaryu2018skills, chioda2021making, barrera2023hard}. The benefits of soft-skills training have additionally been measured through social outcomes that correlate with general quality of life~\citep{oreopoulos2011priceless, heckman2012hard}.

We believe NLP researchers will develop more impactful systems by taking both intrinsic and extrinsic evaluation perspectives. As specific methods begin to show promise through intrinsic automated metrics, they can begin to be utilized as part of human-centered systems design processes. Once a particular system is deemed satisfactory and capable by initial users, it can begin real-world trials to assess impacts from long-term use. This form of development---where deployment is staged alongside increasingly broad evaluation criteria---is key to the safe rollout of research with both high impact and potential risk~\citep{mohs2017drug}. For APAM and other NLP systems where value is derived through direct user interaction, we benefit by learning best practices specific to each domain.

Finally, given the high stakes but relatively low-cost of access to these tools, providing algorithmic auditors transparent access to the systems throughout this process should be standard practice in the development of APAM systems~\citep{raji2020closing}. Risk factors and adverse events (e.g., simulation failures, hallucination, over-reliance) stemming from any of these evaluation procedures should be released in detail, rather than reported in aggregate, in order to facilitate external analysis of possible trends such as via the use of best practices in medical trials~\citep{food2009adverse}. 

\section{Discussion}
\subsection{Societal Impact}
The APAM framework can help in a variety of societal applications. Broadly, APAM can be designed to help increase soft skills like self-awareness, social awareness, self-regulation, and relationship building, all of which can lead to personal wellbeing and broader social good outcomes \cite{jagers2018equity}.  Take the soft skill of self-awareness as an example: self-awareness is an understanding of personal and cultural identity, which correlates with physical wellbeing 
\cite{taylor2010know,schwartz2008broadening}, mental health \cite{bhugra2005migration}, and beneficial learning outcomes \cite{altugan2015relationship}. Psychologically, self-awareness is a foundation for optimism, confidence, and a sense of agency. Relatedly, training self-regulation skills like stress management and intrinsic motivation via APAM can lead to healthy lifestyle choices \cite{antoni2006stress}. Relationship building skills like conflict resolution and communication enhance group learning and cooperative work, but most importantly for the individual, these strong relationships are expected to provide critical social support \cite{seeman2001social} and a higher quality of life \cite{cohen2004social}. Additionally, skills of empathy and perspective-taking form a broader social awareness and are the foundations of strong citizenry \cite{wray2011developmental}. Collectively, our APAM framework is expected to provide both individual and societal benefits in a more equitable manner, as it is designed to provide social learning opportunities to everyone.

APAM is a curriculum design tool that could enable learners to co-construct their learning paths could empower people to actively discover, define, and fill new niches in the economy. Some concrete applications of APAM are achievable in the short term. However, the potential impact of this broader vision necessitates further exploration. This requires new experts to design, train, and maintain AI tooling \cite{wilson2017jobs} for APAM, as well as curriculum design experts to assess when and where true practice and mentorship is irreplaceable. 

Even in areas where current systems for social learning of domain-critical soft skills exist, they are often inaccessible, especially to the socioeconomically disadvantaged. Beyond improving the quality of existing outcomes, APAM could make the same high-quality learning environments available to everyone. AI alone is unlikely to solve all educational and social inequity. However, by focusing on skills that are often learned informally within social groups, APAM can reduce structural inequities which often compound for the already disadvantaged across generations \cite{oded2011inequality}.

\subsection{Concerns and Mitigation}
Despite APAM's benefits, there are still a set of issues we need to be aware of when building systems that can train social skills: 

\paragraph{Stereotypes.} LLM simulations can output caricatures \cite{cheng2023compost} when prompted with broad characteristics such as race and gender. Often, stereotypes arise from under-description in the prompt (e.g., stereotypically casting a "boss" as a white male).  We recommend that system designers highlight and encourage users to specify attributes that are important to the simulation (e.g., gender), enabling users to make changes as needed in the interface. Existing metrics of caricature can be used to raise warnings to users when simulations drift towards stereotypes while giving users full control over the personas with which they practice.

\paragraph{Distributional Shifts.} APAM is designed primarily as a safe training environment in which users can build, practice, and transfer their social skills to the real world. We recommend that system designers should identify and clearly communicate the limitations of the simulation and its deviations from reality. We also recommend that any system based on APAM should take a human-centered development process, observing and interviewing real users to understand the system's behaviors and gather feedback for iteration. Such evaluation will help track the nature and extent of any discrepancies between the simulation and the real world. Finally, to guard against users who might rely on the system too heavily, we recommend that the system caution users against overuse.

\paragraph{Job Risks.} APAM is not designed as a direct replacement for paid human labor. Instead, APAM is a curriculum design tool that allows learners to co-construct their learning paths, empowering people to actively discover, define, and fill niches in the economy. At the corporate level, social skill training programs will still need professional supervision, and these professionals can use automated tools for training events, just as they might have used a static curriculum or textbook in the past. Some individuals may opt for a cheap or free standalone option if they are on the margin. As such, we weigh this risk against the potential benefit of a free-to-use tool, which can assist a broader user population, especially those without professional training or social capital. Professional experts will be able to focus on more tailored, challenging scenarios for skill training, while still maintaining their high level of expertise and uniqueness.

\section{Summary and Outlook}
This perspective paper examines a widespread challenge: mastering essential social skills for both personal and professional success. Opportunities to practice these skills in a safe learning environment are limited, especially for underprivileged groups. We show how LLMs can help create environments where everyone can practice social skills via our proposed AI Partner and AI Mentor framework. Here, the AI Partner offers a risk-free practice  environment, while the AI Mentor provides knowledgeable tailored advice. 

Below we highlight a few take-aways that illustrate how this approach can reshape social skills learning moving forward.  Firstly, utilizing LLMs on APAM requires addressing multiple technical challenges, such as enhancing the simulation of AI partners to exhibit a consistent, plausible and instructive personality, and building AI mentors to have context awareness, domain expertise and feedback efficiency.  Secondly, deploying LLM based social skill training systems has the potential to amplify limitations such as hallucinations and biases, thus our APAM framework offers a roadmap for how to use LLMs for social skill training by breaking safe usage into a continuum dependent on current capabilities. That is, the safe deployment of APAM should emphasize a gradual, risk-aware approach, as controllability and consistency improve for LLMs. Additionally, training social skills via LLMs might suffer from stereotypes and biases in LLM based simulation, distribution shifts and user reliance, as well as potential risks around job replacement. We recommend system designers take a human-centered development process, together with formative evaluations that iterate on the system's behaviors using feedback from observation sessions with real users.  

Overall, the success of APAM depends on fostering interdisciplinary collaborations and team science across diverse research fields and across both academic and professional communities. Such a balanced, intentional, and collaborative process is essential for using LLMs for social good, particularly in areas such as social skill training.

\bibliography{ref}

\end{document}